\documentclass[conference]{IEEEtran}
\IEEEoverridecommandlockouts

\usepackage{cite}
\usepackage{amsmath,amssymb,amsfonts}
\usepackage{algorithmic}
\usepackage{graphicx}
\usepackage{textcomp}
\usepackage{xcolor}
\usepackage{lipsum}
\usepackage{booktabs}
\usepackage[table]{xcolor}
\usepackage{lipsum}
\usepackage{comment}
\usepackage{hyperref}
\usepackage{url}
\usepackage{enumitem}
\usepackage{subcaption}

\newcommand{\resultnof}[2]{#1 \footnotesize $\pm$ #2}

\def\BibTeX{{\rm B\kern-.05em{\sc i\kern-.025em b}\kern-.08em
    T\kern-.1667em\lower.7ex\hbox{E}\kern-.125emX}}
\begin{document}

\title{FedCMAPSS: A Benchmark for Federated Learning in Remaining Useful Life Estimation
}

\author{\IEEEauthorblockN{Amelia Sorrenti}
\IEEEauthorblockA{\textit{University of Catania, Italy}\\
amelia.sorrenti@unict.it}
\and
\IEEEauthorblockN{Matteo Pennisi}
\IEEEauthorblockA{\textit{University of Catania, Italy} \\
matteo.pennisi@unict.it}
\and
\IEEEauthorblockN{Concetto Spampinato}
\IEEEauthorblockA{\textit{University of Catania, Italy} \\
concetto.spampinato@unict.it}
\and
\IEEEauthorblockN{Simone Palazzo}
\IEEEauthorblockA{\textit{University of Catania, Italy} \\
simone.palazzo@unict.it}
}

\maketitle

\begin{abstract}
Data-driven prognostics and health management has emerged as a key enabler for Industry 4.0, yet the development of robust remaining useful life (RUL) estimation models is often limited by the scarcity of run-to-failure data. While federated learning offers a promising paradigm to collaboratively train predictive models without sharing sensor data, research efforts have operated so far in the absence of a common evaluation framework. To address this gap, this paper introduces FedCMAPSS, a benchmark for federated RUL estimation based on the commonly-used NASA C-MAPSS dataset. We define a set of five standardized tasks designed to simulate real-world industrial challenges, ranging from ideal IID settings to extreme statistical heterogeneity, and conduct a systematic evaluation of state-of-the-art federated optimization algorithms across multiple neural architectures. By establishing reproducible baselines and making the source code and data splits publicly available, this work aims to provide a standard foundation for developing and comparing federated predictive maintenance solutions.
\end{abstract}

\begin{IEEEkeywords}
federated learning, predictive maintenance, remaining useful life
\end{IEEEkeywords}

\section{Introduction}
In modern industrial systems, predictive maintenance represents a critical shift from reactive and preventative strategies to data-driven prognostics and health management (PHM)~\cite{vachtsevanos2006intelligent}, with remaining useful life (RUL) estimation standing out as one of the primary prognostic objectives~\cite{hanif2018comprehensive}. In this context, deep neural networks have increasingly become the standard in PHM applications, leveraging their capacity to automatically extract high-level representations from multidimensional sensor data~\cite{zhang2019review,rezaeianjouybari2020deep,polverino2023machine}.

Despite their success, these architectures require large amounts of centralized run-to-failure data for optimization, while organizations often lack sufficient samples to train robust generalized regression models, especially in real-world edge deployments~\cite{qin2023dynamic}; at the same time, they may not be willing to share their data with other stakeholders, due to intellectual property issues or data privacy laws (such as the GDPR)~\cite{arunan2023federated}. A possible solution to these problems may lie in the adoption of federated learning (FL) techniques, which enable distributed nodes to collaboratively optimize a global model by transmitting only gradient updates or network weights rather than raw data~\cite{mcmahan2017communication,yang2019federated,pennisi2022gan,mineo2023fedetr}. The benefits of this solution have led to significant research efforts in the field, but the challenge is still open: implementing federated learning in resource-constrained IoT environments introduces bandwidth and computational constraints~\cite{nguyen2021federated}, exacerbated by the communication overhead associated with transmitting deep neural networks~\cite{chen2023bearing} and by security aspects~\cite{bonawitz2017practical,zhu2024collaborative}.

The convergence of federated optimization protocols and continuous RUL estimation is a promising direction for scaling prognostic solutions across distributed edge devices. Recent studies validate that decentralized neural networks optimized via federated topologies can achieve predictive accuracy comparable to centralized models~\cite{barbosa2025using,pennisi2023experience}. However, complex and non-stationary degradation signals present unique optimization challenges for distributed learning architectures: for instance, standard aggregation protocols like FedAvg~\cite{mcmahan2017communication} exhibit learning oscillations and high uncertainty during the optimization of multilayer perceptrons across heterogeneous sensor data distributions~\cite{rosero2020remaining}.

A significant obstacle to the advancement of federated RUL estimation is the lack of consistency in evaluation settings and data partitioning strategies across existing literature. For instance, many works evaluate data-driven RUL algorithms on the NASA Commercial Modular Aero-Propulsion System Simulation (C-MAPSS) dataset~\cite{saxena2008damage,frederick2007user}, modeling damage propagation in aircraft turbofan engines. However, even on the same dataset, the research community often utilizes inconsistent evaluation protocols, making it difficult to fairly compare competing prognostic algorithms~\cite{saxena2008metrics,byington2005verification}. For instance, the lack of consistency on how data samples are distributed among clients (and on the number of clients itself) can drastically affect evaluation variability, especially because of the nature of the dataset, which features samples deployed in significantly different operating conditions. As a result, this statistical heterogeneity makes cross-paper comparisons impossible.

To address these systemic evaluation discrepancies, this paper introduces a standardized benchmark for FL applied to RUL estimation. Based on the C-MAPSS dataset, we define a set of standard tasks supporting the evaluation of FL methods in different data distribution settings, aligned to predictive maintenance contexts and objectives. We also provide a thorough comparison between state-of-the-art FL algorithms and neural network architectures employed for RUL estimation. Overall, our goal is to provide the prognostic machine learning community with a common dataset and a reproducible framework, to accurately evaluate decentralized predictive maintenance algorithms. All code for reproducing and extending the results of the paper is provided at \texttt{\url{https://github.com/perceivelab/FedCMAPSS}}.

\section{Related Work}
{Predictive maintenance} has evolved from traditional analytical models to data-driven diagnostic systems designed to detect and classify machinery faults, with deep learning architectures becoming central to these systems due to their ability to extract high-level representations from complex multidimensional sensor streams~\cite{zhang2021federated,liu2020deep,konevcny2016federated}.
RUL estimation extends beyond classification to predict continuous degradation trajectories, requiring models capable of capturing long-term temporal dependencies. Foundational approaches adapted convolutional neural networks to replace manual feature engineering~\cite{sateesh2016deep}. Subsequently, LSTM-based recurrent architectures became the standard for RUL estimation due to their ability to prevent gradient dispersion over long sequences~\cite{hochreiter1997long,zheng2017long,vaccaro2023remaining}. To address the limitation of LSTMs heavily weighing only the final time step, self-attention mechanisms, including transformer architectures, have been introduced to dynamically assign importance weights across the entire time window~\cite{chen2020machine}.

{Federated RUL estimation} has typically established FedAvg as the baseline for aggregating local model parameters centrally without sharing raw data~\cite{mcmahan2017communication}. This standard aggregation has been applied to various deep learning structures, from baseline multilayer perceptrons to recurrent networks~\cite{chen2023remaining,barbosa2025using}. However, since coordinate-wise averaging may suffer from destructive interference between model parameters, more complex matching algorithms~\cite{wang2020federated} have been adapted to RUL estimation~\cite{arunan2023federated}. FedProx introduces a proximal term to restrict local optimization updates, reducing learning oscillations~\cite{li2020federated,rosero2020remaining}. Other approaches combine this constraint with a shared-data strategy to stabilize training across non-IID (independent and identically distributed) data~\cite{lai2024fedcbe}. Dynamic weighting mechanisms adjust the global aggregation based on client quality~\cite{qin2023dynamic,zhu2024collaborative}. Adaptive client momentum approaches compute local momentum to outperform traditional server-side momentum aggregation~\cite{yilmaz2025federated,sun2024role}. 
Federated personalization addresses non-IID data by partitioning models into global base layers and local personalization layers, effectively applied to 1D CNNs for RUL prediction~\cite{arivazhagan2019federated,soderkvist2024collaborative}.
To reduce communication overhead in complex topologies, Fed-TNP employs model pruning to sparsify network weights before transmission~\cite{chen2023bearing}.

In terms of {evaluation settings}, a discrepancy among studies lies in the partitioning of client data. Some approaches utilize homogeneous splits, leveraging random partitioning to simulate distributed environments~\cite{kamei2023comparison}. Conversely, other research applies non-IID partitioning, which complicates the training process compared to homogeneous splits~\cite{arunan2023federated}. Furthermore, preprocessing and target normalization strategies vary widely across the literature. While some works rely on simple min-max normalization across the entire dataset~\cite{kamei2023comparison}, others employ clustering to identify operating regimes prior to applying Z-score standardizations~\cite{peel2008data,rosero2020remaining}. Target label generation also exhibits variability, where piecewise linear rectification strategies are commonly employed with different capping values~\cite{soderkvist2024collaborative,kamei2023comparison}.

\section{Method}

\subsection{Dataset}
We build our federated RUL estimation benchmark upon the widely-used C-MAPSS dataset~\cite{saxena2008damage}. The dataset simulates the degradation of aircraft turbofan engines over time, capturing multidimensional sensor readings under various operational states and fault conditions. It is divided into four distinct sub-datasets, namely FD001, FD002, FD003, and FD004, each presenting varying degrees of complexity. Specifically, FD001 and FD003 simulate a single operating condition at sea level, whereas FD002 and FD004 encompass six different operational settings, increasing the statistical heterogeneity of the sensor data. Furthermore, the datasets model different degradation mechanisms: FD001 and FD002 feature a single fault mode related to high-pressure compressor degradation, while FD003 and FD004 introduce a second concurrent fault mode affecting the engine fan. The volume of data also varies across the subsets, with FD001 and FD003 containing 100 training and 100 testing trajectories each, FD002 containing 260 training and 259 testing trajectories, and FD004 containing 248 training and 249 testing trajectories.

Each trajectory represents a multivariate time series extracted from a single engine. Raw data describe 26 features, which include the engine unit identifier, the elapsed time expressed in cycles, three operational settings, and 21 distinct sensor measurements contaminated with realistic sensor noise. At the beginning of each time series, the engine operates normally, possessing an unknown degree of initial wear that does not constitute a fault condition. In the training subsets, this degradation grows until system failure; conversely, the test trajectories are truncated at a random point prior to failure.

\subsection{Tasks}

We organize the C-MAPSS dataset into five \emph{tasks}, designed to model real-world industrial challenges.
Each task defines a fixed number of clients, among which dataset samples are distributed, based on the task's specific objective. To ensure the statistical validity and reproducibility of the benchmark, we generate 10 independent random splits for each task. Within each split, the data allocated to every client is partitioned into a \emph{training} subset, utilized for local model optimization; a \emph{testing} subset, consisting of the standard truncated C-MAPSS test trajectories and used to compute the global evaluation metrics; and a \emph{full testing} subset, which holds unused training samples containing complete run-to-failure trajectories, reserved for generating plots of actual RUL versus predicted RUL. Except where constrained by specific task settings, we generally apply an 85\% subsampling ratio to allocate trajectories to the training subset, reserving the remaining 15\% for the full testing subset. In detail, tasks are defined as follows:
\begin{itemize}[leftmargin=*]
\item \textbf{Task A: IID baseline.} This task assesses convergence in an ideal scenario. We utilize the FD001 dataset, distributing engine trajectories among 10 clients following an IID partitioning strategy. No domain shift or label skew is introduced between clients.
\item \textbf{Task B: domain shift.} This task emulates cross-silo collaboration, where client data distributions reflect different mechanical paradigms or fault modes. We leverage all four C-MAPSS sub-datasets and assign each to one client. 
\item \textbf{Task C: label skew.} This task introduces label skew, with each client observing different lifecycle lengths. We utilize the FD004 dataset (with six operational conditions and two fault models) and sort units by their total lifespan. The sorted trajectories are then sequentially assigned to 10 clients, forcing them to train exclusively on engines with a specific lifespan range.
\item \textbf{Task D: feature skew.} This task assigns skewed sensor measurement distributions among 6 clients. Using the FD004 dataset, we apply k-means clustering to the operational settings features to isolate the data by condition. Each of the 6 clients is then exclusively assigned data corresponding to one specific operational cluster. Note that this implies that the same engine can be assigned to multiple clients; however, each of them will ``see'' only the cycles in which the engine operated at the client's assigned condition, potentially introducing temporal gaps in the time series.
\item \textbf{Task E: few-shot learning.} This task scales the network to 85 clients using the FD001 dataset, assigning one unique training trajectory to each client, thus forcing the aggregation of model parameters from highly biased local datasets.
\end{itemize}

\subsection{Architectures}
\label{sec:architectures}
We evaluate representative neural architectures for federated RUL prediction, inspired by recent literature, including recurrent, convolutional, and attention-based models. Implementation details can be found in the source code repository. The architectures under comparison are:
\begin{itemize}[leftmargin=*]
    \item \textbf{LSTM}~\cite{arunan2023federated}: a simple LSTM network with a single layer.
    \item \textbf{RNN}~\cite{chen2023remaining}: a five-layer RNN progressively reducing hidden dimensionality, followed by a multi-layer perceptron (MLP) for prediction.
    \item \textbf{CNN}~\cite{chen2023federated}: a convolutional architecture treating a multivariate time series as a 2D matrix with asymmetric kernels, followed by an MLP for RUL estimation.
    \item \textbf{AFT}~\cite{zhu2024collaborative}: an Attention-Free Transformer (AFT) processes the input sequence, before feeding it to two 2D convolutional layers, followed by an MLP for prediction.
    \item \textbf{AttBiGRU}~\cite{qin2023dynamic}: a bidirectional GRU is followed by 4-head self-attention to dynamically weigh the contribution of different signal parts; the resulting attention map is flattened and processed by an MLP with dropout.
\end{itemize}
Model hyperparameters are set to default values from the corresponding papers, when available, or following established heuristics.

\subsection{Federated methods}
\label{sec:methods}
We benchmark different federated optimization methods, including classical baselines and more recent approaches designed for non-IID federated training:
\begin{itemize}[leftmargin=*]
\item \textbf{FedAvg}~\cite{mcmahan2017communication} is the classical FL baseline. At each communication round, clients train the shared model locally and the server aggregates client updates through weighted averaging, typically proportional to local sample sizes. 
\item \textbf{SCAFFOLD}~\cite{karimireddy2020scaffold} addresses \emph{client drift} caused by data heterogeneity using control variates maintained by the server and each client, improving convergence speed and final accuracy in non-IID settings.
\item \textbf{FedDyn}~\cite{acar2021federated} introduces a \emph{dynamic regularization} objective to align local optima with the global objective, especially under non-IID data. 
\item \textbf{FedCross}~\cite{hu2024fedcross} proposes a multi-model cross-aggregation strategy rather than relying on a single global model during each round, using multiple middleware models. 
\end{itemize}
Method-specific hyperparameters for FL methods are set to default values in the PFLLib\footnote{\texttt{\url{https://www.pfllib.com}}} library.

\subsection{Evaluation metrics}
We evaluate RUL prediction performance using the \textit{Root Mean Square Error} (RMSE) and the \textit{NASA Score}, which are both standard in prognostics benchmarks.

Given ground-truth RUL values $y_i$ and predictions $\hat{y}_i$ for $N$ test samples, the RMSE is defined as
\begin{equation}
\mathrm{RMSE} = \sqrt{\frac{1}{N}\sum_{i=1}^{N}(y_i-\hat{y}_i)^2 }.
\end{equation}
RMSE measures the average magnitude of prediction errors and is easy to interpret in the same unit as the target RUL.

In addition, we report the NASA Score, an asymmetric metric that penalizes late predictions (i.e., overestimation of RUL) more severely than early predictions. Let $\Delta_i = \hat{y}_i - y_i$. The score is computed as
\begin{equation}
\mathrm{Score} = \sum_{i=1}^{N} s_i,
\quad
s_i =
\begin{cases}
\exp\!\left(-\frac{\Delta_i}{13}\right)-1, & \Delta_i < 0,\\[4pt]
\exp\!\left(\frac{\Delta_i}{10}\right)-1, & \Delta_i \ge 0.
\end{cases}
\end{equation}
Lower values indicate better performance for both metrics. The asymmetric NASA Score is particularly relevant in maintenance applications, where overly optimistic RUL estimates may lead to delayed interventions. The NASA Score is commonly reported cumulatively over all test samples.

\begin{table*}[htbp]
    \caption{Local client performance in the isolated setting (RMSE).}
    \renewcommand{\arraystretch}{1.2}
    \centering
    \begin{tabular}{l|ccccc}
        \toprule
        \textbf{}     & \textbf{Task A} & \textbf{Task B} & \textbf{Task C} & \textbf{Task D} & \textbf{Task E}\\
        \midrule
        \rowcolor{gray!15}
        \textbf{LSTM} & \resultnof{21.53}{0.57} & \resultnof{21.27}{0.14} & \resultnof{28.66}{0.20} & \resultnof{22.58}{0.09} & \resultnof{37.16}{0.59}\\
        \textbf{AFT} & \resultnof{40.40}{0.14} & \resultnof{37.26}{0.38} & \resultnof{46.14}{0.03} & \resultnof{39.93}{0.03} & \resultnof{42.08}{0.58}\\
        \rowcolor{gray!15}
        \textbf{AttBiGRU} & \resultnof{25.47}{0.88} & \resultnof{22.37}{0.19} & \resultnof{23.04}{0.36} & \resultnof{33.05}{0.07} & \resultnof{36.94}{0.61}\\
        \textbf{RNN} & \resultnof{40.34}{0.10} & \resultnof{43.29}{0.04} & \resultnof{45.27}{0.03} & \resultnof{39.82}{0.02} & \resultnof{41.22}{0.51}\\
        \rowcolor{gray!15}
        \textbf{CNN} & \resultnof{40.88}{0.21} & \resultnof{37.77}{0.51} & \resultnof{46.47}{0.04} & \resultnof{39.37}{0.22} & \resultnof{40.82}{0.47}\\
        \bottomrule
        \end{tabular}
        \label{tab:task_local}
\end{table*}

\begin{table*}[htbp]
    \caption{Results on Task A by architecture and federated method.}
    \renewcommand{\arraystretch}{1.2}
    \centering
        \begin{tabular}{l|cc|cc|cc|cc}
        \toprule
        \textbf{}     & \multicolumn{2}{c}{\textbf{FedAvg}} & \multicolumn{2}{c}{\textbf{SCAFFOLD}} & \multicolumn{2}{c}{\textbf{FedDyn}} & \multicolumn{2}{c}{\textbf{FedCross}}\\
        \cmidrule(lr){2-3} \cmidrule(lr){4-5} \cmidrule(lr){6-7} \cmidrule(lr){8-9}
        \textbf{} & \textbf{RMSE} & \textbf{NASA$\times 10^{-3}$} & \textbf{RMSE} & \textbf{NASA$\times 10^{-3}$} & \textbf{RMSE}& \textbf{NASA$\times 10^{-3}$} & \textbf{RMSE}& \textbf{NASA$\times 10^{-3}$}\\
        \cmidrule(lr){1-1} \cmidrule(lr){2-3} \cmidrule(lr){4-5} \cmidrule(lr){6-7} \cmidrule(lr){8-9}

        \rowcolor{gray!15}
        \textbf{LSTM}
        & \resultnof{17.66}{1.11} & \resultnof{0.99}{0.43}
        & \resultnof{17.36}{1.11} & \resultnof{0.81}{0.27}
        & \resultnof{23.45}{0.72} & \resultnof{2.38}{0.63}
        & \resultnof{18.81}{1.08} & \resultnof{1.06}{0.53}\\

        \textbf{AFT}
        & \resultnof{17.19}{1.09} & \resultnof{0.81}{0.26}
        & \resultnof{16.59}{1.41} & \resultnof{0.88}{0.49}
        & \resultnof{43.71}{1.12} & \resultnof{9.78}{1.86}
        & \resultnof{18.79}{1.30} & \resultnof{1.47}{0.93}\\

        \rowcolor{gray!15}
        \textbf{AttBiGRU}
        & \resultnof{17.84}{1.01} & \resultnof{1.14}{0.76}
        & \resultnof{17.46}{1.67} & \resultnof{0.92}{0.45}
        & \resultnof{31.32}{3.17} & \resultnof{12.79}{6.43}
        & \resultnof{22.37}{1.59} & \resultnof{2.95}{3.72}\\

        \textbf{RNN}
        & \resultnof{18.97}{0.81} & \resultnof{1.10}{0.28}
        & \resultnof{18.96}{0.88} & \resultnof{1.11}{0.29}
        & \resultnof{41.82}{0.31} & \resultnof{6.89}{0.42}
        & \resultnof{20.01}{0.80} & \resultnof{1.34}{0.41}\\

        \rowcolor{gray!15}
        \textbf{CNN}
        & \resultnof{14.79}{0.58} & \resultnof{0.39}{0.14}
        & \resultnof{14.56}{0.55} & \resultnof{0.40}{0.13}
        & \resultnof{40.45}{0.41} & \resultnof{6.62}{1.33}
        & \resultnof{15.48}{0.76} & \resultnof{0.39}{0.07}\\

        \bottomrule
        \end{tabular}
        \label{tab:task_a}
\end{table*}

\subsection{Evaluation protocol}

Following common RUL estimation practice, sensor signals are split into fixed-length temporal windows and paired with the corresponding RUL labels. We apply client-side preprocessing only (e.g., normalization and windowing) before local training. During training, local optimization aims to minimize the mean squared error (MSE) loss between the predicted and ground-truth RUL. This objective may be augmented by any loss terms introduced by the respective FL algorithms. Following the piecewise linear degradation assumption, widely adopted in C-MAPSS literature, the continuous target RUL values are clipped to a maximum number of cycles, set to 125, and linearly normalized to the $[0,1]$ interval.

Model training is carried out using standard stochastic gradient descent; relevant hyperparameters (learning rate, batch size) are tuned for each model architecture by conducting a grid search using a centralized training paradigm on the FD001 dataset. We currently do not tune optimization hyperparameters in the federated setting, nor model-specific or algorithm-specific hyperparameters. 
In the federated setting, we train all algorithms for 100 rounds, varying the number of epochs of local training per round (1, 5 and 10).

Our experimental framework omits a dedicated validation set during the federated evaluation phase, and unconditionally evaluates the global model aggregated at the final communication round. This is chosen to maximize procedural simplicity and reflect realistic edge deployment constraints, where auxiliary validation data is often unavailable. Moreover, empirical observations from the initial centralized hyperparameter selection showed that the optimization trajectory is generally smooth, exhibiting minimal oscillatory behavior across epochs. Nevertheless, researchers utilizing our benchmark who wish to employ alternative model selection criteria, such as early stopping, should independently partition the local training sets to construct validation splits, to ensure methodological rigor. 

Finally, to ensure statistical significance, we evaluate all combinations of model and FL algorithm on all 10 independent random splits generated for each task, reporting mean and standard deviation of both the RMSE and the NASA Score on the test set.

\section{Experimental Results}

We evaluate the proposed benchmark across the five defined tasks. We first evaluate isolated local training and full-data centralized optimization. We then present a comparative analysis of model architectures and FL algorithms. Finally, we analyze the optimization dynamics concerning communication overhead and provide a qualitative assessment of the predicted degradation trajectories.

\subsection{Local baseline}

To contextualize the performance of the FL methods, we establish an empirical lower-bound baseline under a local-only training regime, where each client updates its model using only its isolated partition without any parameter synchronization.

Results for all tasks and model architectures are reported in Tab.~\ref{tab:task_local}
in terms of RMSE.
Averaging results across architectures within each task highlights a pronounced difficulty gap in the isolated setting. Overall, however, all models struggle when tackling each task locally. 

\begin{table*}[htbp]
    \caption{Results on Task B by architecture and federated method.}
    \renewcommand{\arraystretch}{1.2}
    \centering
        \begin{tabular}{l|cc|cc|cc|cc}
        \toprule
        \textbf{}     & \multicolumn{2}{c}{\textbf{FedAvg}} & \multicolumn{2}{c}{\textbf{SCAFFOLD}} & \multicolumn{2}{c}{\textbf{FedDyn}} & \multicolumn{2}{c}{\textbf{FedCross}}\\
        \cmidrule(lr){2-3} \cmidrule(lr){4-5} \cmidrule(lr){6-7} \cmidrule(lr){8-9}
        \textbf{} & \textbf{RMSE} & \textbf{NASA$\times 10^{-3}$} & \textbf{RMSE} & \textbf{NASA$\times 10^{-3}$} & \textbf{RMSE}& \textbf{NASA$\times 10^{-3}$} & \textbf{RMSE}& \textbf{NASA$\times 10^{-3}$}\\
        \cmidrule(lr){1-1} \cmidrule(lr){2-3} \cmidrule(lr){4-5} \cmidrule(lr){6-7} \cmidrule(lr){8-9}

        \rowcolor{gray!15}
        \textbf{LSTM} 
        & \resultnof{21.14}{0.58} & \resultnof{9.85}{2.20} 
        & \resultnof{18.42}{0.38} & \resultnof{7.06}{0.79} 
        & \resultnof{40.83}{1.61} & $\sigma \approx 36$ 
        & \resultnof{16.43}{0.31} & \resultnof{5.35}{1.25}\\

        \textbf{AFT} 
        & \resultnof{29.30}{3.22} & $\sigma \approx 185$ 
        & \resultnof{17.30}{0.86} & \resultnof{6.11}{1.81} 
        & \resultnof{47.17}{3.19} & $\sigma \approx 60$ 
        & \resultnof{15.85}{0.63} & \resultnof{3.87}{0.79}\\

        \rowcolor{gray!15}
        \textbf{AttBiGRU} 
        & \resultnof{27.17}{1.35} & $\sigma \approx 40$ 
        & \resultnof{17.84}{0.37} & \resultnof{5.96}{0.56} 
        & \resultnof{38.24}{2.32} & $\sigma \approx 530$ 
        & \resultnof{16.90}{0.38} & \resultnof{4.49}{0.79}\\

        \textbf{RNN} 
        & \resultnof{28.03}{2.74} & $\sigma \approx 87$ 
        & \resultnof{20.06}{0.37} & \resultnof{11.99}{1.92} 
        & \resultnof{44.82}{0.35} & \resultnof{67.59}{4.09} 
        & \resultnof{20.27}{0.51} & \resultnof{10.45}{2.88}\\

        \rowcolor{gray!15}
        \textbf{CNN} 
        & \resultnof{46.78}{3.02} & $\sigma \approx 370$ 
        & \resultnof{37.30}{1.71} & $\sigma \approx 86$ 
        & \resultnof{42.26}{0.34} & \resultnof{49.42}{6.79} 
        & \resultnof{33.69}{1.31} & $\sigma \approx 206$ \\

        \bottomrule
        \end{tabular}
        \label{tab:task_b}
\end{table*}

\begin{table*}[htbp]
    \caption{Results on Task C by architecture and federated method.}
    \renewcommand{\arraystretch}{1.2}
    \centering
        \begin{tabular}{l|cc|cc|cc|cc}
        \toprule
        \textbf{}     & \multicolumn{2}{c}{\textbf{FedAvg}} & \multicolumn{2}{c}{\textbf{SCAFFOLD}} & \multicolumn{2}{c}{\textbf{FedDyn}} & \multicolumn{2}{c}{\textbf{FedCross}}\\
        \cmidrule(lr){2-3} \cmidrule(lr){4-5} \cmidrule(lr){6-7} \cmidrule(lr){8-9}
        \textbf{} & \textbf{RMSE} & \textbf{NASA$\times 10^{-3}$} & \textbf{RMSE} & \textbf{NASA$\times 10^{-3}$} & \textbf{RMSE}& \textbf{NASA$\times 10^{-3}$} & \textbf{RMSE}& \textbf{NASA$\times 10^{-3}$}\\
        \cmidrule(lr){1-1} \cmidrule(lr){2-3} \cmidrule(lr){4-5} \cmidrule(lr){6-7} \cmidrule(lr){8-9}

        \rowcolor{gray!15}
        \textbf{LSTM}
        & \resultnof{29.48}{0.34} & \resultnof{38.26}{4.96}
        & \resultnof{27.31}{1.05} & \resultnof{31.73}{21.26}
        & \resultnof{39.92}{2.40} & \resultnof{15.01}{4.61}
        & \resultnof{17.28}{0.92} & \resultnof{1.44}{0.96}\\

        \textbf{AFT}
        & \resultnof{30.32}{0.33} & \resultnof{41.59}{5.97}
        & \resultnof{31.22}{1.91} & $\sigma \approx 55$ 
        & \resultnof{49.44}{1.26} & \resultnof{48.12}{8.59}
        & \resultnof{17.75}{0.70} & \resultnof{1.35}{0.62}\\

        \rowcolor{gray!15}
        \textbf{AttBiGRU}
        & \resultnof{30.87}{0.24} & \resultnof{27.14}{7.59}
        & \resultnof{32.15}{2.39} & $\sigma \approx 50$ 
        & \resultnof{36.30}{2.11} & $\sigma \approx 95$ 
        & \resultnof{20.18}{0.68} & \resultnof{1.89}{0.60}\\

        \textbf{RNN}
        & \resultnof{30.52}{0.50} & \resultnof{27.10}{5.97}
        & \resultnof{34.69}{4.51} & $\sigma \approx 41$ 
        & \resultnof{45.24}{0.28} & \resultnof{23.75}{1.27}
        & \resultnof{18.24}{0.33} & \resultnof{1.32}{0.34}\\

        \rowcolor{gray!15}
        \textbf{CNN}
        & \resultnof{35.12}{0.38} & $\sigma \approx 39$ 
        & \resultnof{33.10}{0.59} & $\sigma \approx 58$ 
        & \resultnof{42.93}{0.20} & \resultnof{17.76}{2.79}
        & \resultnof{26.22}{0.68} & $\sigma \approx 55$\\ 

        \bottomrule
        \end{tabular}
        \label{tab:task_c}
\end{table*}

\begin{table*}[htbp]
    \caption{Results on Task D by architecture and federated method.}
    \renewcommand{\arraystretch}{1.2}
    \centering
        \begin{tabular}{l|cc|cc|cc|cc}
        \toprule
        \textbf{}     & \multicolumn{2}{c}{\textbf{FedAvg}} & \multicolumn{2}{c}{\textbf{SCAFFOLD}} & \multicolumn{2}{c}{\textbf{FedDyn}} & \multicolumn{2}{c}{\textbf{FedCross}}\\
        \cmidrule(lr){2-3} \cmidrule(lr){4-5} \cmidrule(lr){6-7} \cmidrule(lr){8-9}
        \textbf{} & \textbf{RMSE} & \textbf{NASA$\times 10^{-3}$} & \textbf{RMSE} & \textbf{NASA$\times 10^{-3}$} & \textbf{RMSE}& \textbf{NASA$\times 10^{-3}$} & \textbf{RMSE}& \textbf{NASA$\times 10^{-3}$}\\
        \cmidrule(lr){1-1} \cmidrule(lr){2-3} \cmidrule(lr){4-5} \cmidrule(lr){6-7} \cmidrule(lr){8-9}

        \rowcolor{gray!15}
        \textbf{LSTM}
        & \resultnof{14.12}{0.65} & \resultnof{4.23}{0.61}
        & \resultnof{12.58}{0.25} & \resultnof{3.85}{0.21}
        & \resultnof{24.67}{1.05} & \resultnof{14.51}{4.28}
        & \resultnof{12.28}{0.23} & \resultnof{3.52}{0.31}\\

        \textbf{AFT}
        & \resultnof{16.30}{0.64} & $\sigma \approx 90$ 
        & \resultnof{15.64}{0.21} & $\sigma \approx 17$ 
        & \resultnof{39.86}{0.38} & $\sigma \approx 20$ 
        & \resultnof{15.44}{0.41} & $\sigma \approx 67$\\ 

        \rowcolor{gray!15}
        \textbf{AttBiGRU}
        & \resultnof{15.15}{0.25} & \resultnof{4.89}{0.57}
        & \resultnof{14.13}{0.08} & \resultnof{5.20}{0.50}
        & \resultnof{33.84}{0.93} & $\sigma \approx 26$ 
        & \resultnof{14.05}{0.15} & \resultnof{4.67}{0.40}\\

        \textbf{RNN}
        & \resultnof{13.84}{0.38} & \resultnof{5.38}{2.11}
        & \resultnof{15.63}{0.11} & \resultnof{4.37}{0.14}
        & \resultnof{40.06}{0.33} & \resultnof{67.82}{14.08}
        & \resultnof{15.05}{0.23} & \resultnof{4.13}{0.21}\\

        \rowcolor{gray!15}
        \textbf{CNN}
        & \resultnof{17.15}{0.31} & \resultnof{6.55}{1.40}
        & \resultnof{17.90}{0.38} & \resultnof{3.90}{0.34}
        & \resultnof{40.24}{0.71} & \resultnof{73.70}{30.43}
        & \resultnof{15.40}{0.22} & \resultnof{3.72}{0.34}\\

        \bottomrule
        \end{tabular}
        \label{tab:task_d}
\end{table*}

\begin{table*}[htbp]
    \caption{Results on Task E by architecture and federated method.}
    \renewcommand{\arraystretch}{1.2}
    \centering
        \begin{tabular}{l|cc|cc|cc|cc}
        \toprule
        \textbf{}     & \multicolumn{2}{c}{\textbf{FedAvg}} & \multicolumn{2}{c}{\textbf{SCAFFOLD}} & \multicolumn{2}{c}{\textbf{FedDyn}} & \multicolumn{2}{c}{\textbf{FedCross}}\\
        \cmidrule(lr){2-3} \cmidrule(lr){4-5} \cmidrule(lr){6-7} \cmidrule(lr){8-9}
        \textbf{} & \textbf{RMSE} & \textbf{NASA$\times 10^{-3}$} & \textbf{RMSE} & \textbf{NASA$\times 10^{-3}$} & \textbf{RMSE}& \textbf{NASA$\times 10^{-3}$} & \textbf{RMSE}& \textbf{NASA$\times 10^{-3}$}\\
        \cmidrule(lr){1-1} \cmidrule(lr){2-3} \cmidrule(lr){4-5} \cmidrule(lr){6-7} \cmidrule(lr){8-9}

        \rowcolor{gray!15}
        \textbf{LSTM}
        & \resultnof{27.30}{0.82} & \resultnof{5.42}{1.01}
        & \resultnof{28.92}{0.99} & \resultnof{9.63}{1.70}
        & \resultnof{34.41}{1.70} & \resultnof{32.86}{11.42}
        & \resultnof{28.15}{1.10} & \resultnof{7.83}{1.73}\\

        \textbf{AFT}
        & \resultnof{34.42}{1.26} & \resultnof{15.70}{3.65}
        & \resultnof{34.81}{1.30} & \resultnof{20.16}{4.52}
        & \resultnof{41.00}{0.45} & \resultnof{6.73}{0.45}
        & \resultnof{36.94}{1.41} & \resultnof{35.02}{6.60}\\

        \rowcolor{gray!15}
        \textbf{AttBiGRU}
        & \resultnof{32.24}{0.80} & \resultnof{8.79}{1.25}
        & \resultnof{32.74}{0.86} & \resultnof{12.47}{1.68}
        & \resultnof{42.29}{3.65} & $\sigma \approx 90$ 
        & \resultnof{33.13}{0.83} & \resultnof{13.10}{2.80}\\

        \textbf{RNN}
        & \resultnof{41.11}{0.86} & \resultnof{5.02}{0.34}
        & \resultnof{40.01}{0.64} & \resultnof{4.31}{0.16}
        & \resultnof{41.10}{0.90} & \resultnof{5.02}{0.37}
        & \resultnof{40.44}{0.74} & \resultnof{4.52}{0.26}\\

        \rowcolor{gray!15}
        \textbf{CNN}
        & \resultnof{29.69}{0.59} & \resultnof{6.10}{1.27}
        & \resultnof{30.35}{1.01} & \resultnof{11.53}{2.92}
        & \resultnof{40.25}{0.51} & \resultnof{5.11}{0.29}
        & \resultnof{32.69}{0.80} & \resultnof{14.47}{3.84}\\

        \bottomrule
        \end{tabular}
        \label{tab:task_e}
\end{table*}

\subsection{Task evaluation}

We then evaluate the model architectures and FL methods introduced in Sect.~\ref{sec:architectures} and \ref{sec:methods}. Results are reported, for each task, in Tab.~\ref{tab:task_a}-\ref{tab:task_e}.

As shown in Tab.~\ref{tab:task_a}, under the IID conditions of Task~A, all methods achieve strong RMSE, with FedAvg and SCAFFOLD performing generally better. Across architectures, LSTM is the only one which does not exhibit method sensitivity, remaining competitive with all FL strategies.

Tab.~\ref{tab:task_b} introduces the difficulties associated with non-IID scenarios, as clients optimize on different operating regimes and fault patterns in Task B. In this setting, drift-mitigation methods tend to be more effective, with FedCross and SCAFFOLD providing the strongest overall performance across architectures. 
An observation should be made here regarding the NASA Score. Since it applies exponential penalties, it is strongly sensitive to even a small number of samples with large prediction errors, leading to cases where the mean and standard deviation are both very large and comparable. Therefore, rather than reporting a meaningless average, in these cases we only show the standard deviation, to give an idea of the degree of score variability. We emphasize that this is a significant limitation of the metric, and more outlier-robust alternatives should be explored in future work. For readability, we report NASA Score values scaled by $10^{-3}$ in all tables.

Continuing with Task~C, a comparable level of difficulty is observed, where clients are exposed to different lifespan ranges (Tab.~\ref{tab:task_c}). Here, FedCross emerges as the most effective approach, and the improvements are most pronounced with the LSTM, AFT and RNN models. 
In contrast, Task~D appears comparatively less critical than the other non-IID regimes (Tab.~\ref{tab:task_d}): despite the distribution shift in the input space, most methods achieve solid performance and the gap among federated strategies narrows, with the notable exception of FedDyn, which exhibits significantly lower prediction accuracy.

Task~E is the most challenging setting, as evidenced by the results in Tab.~\ref{tab:task_e}. Under extremely limited and highly biased local datasets, LSTM and CNN provide the most stable performance among the considered architectures, consistent with reduced sensitivity to local overfitting in this few-shot regime.

\begin{figure*}[ht]
  \centering
  \begin{subfigure}{0.32\textwidth}
    \centering
    \includegraphics[width=\linewidth]{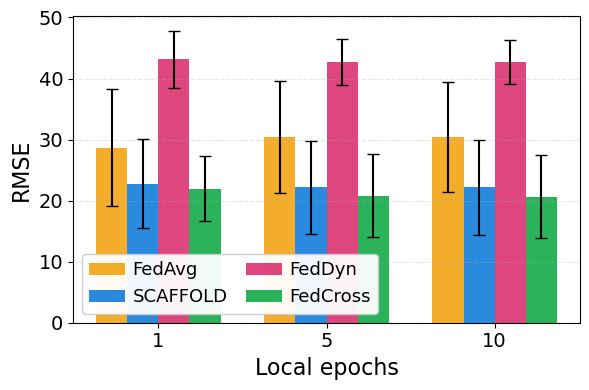}
    \caption{Task B}
  \end{subfigure}\hfill
  \begin{subfigure}{0.32\textwidth}
    \centering
    \includegraphics[width=\linewidth]{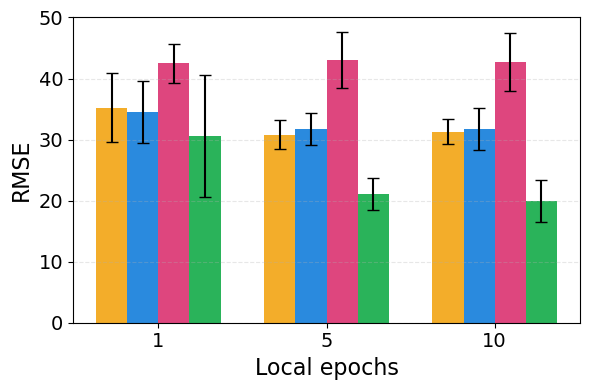}
    \caption{Task C}
  \end{subfigure}\hfill
  \begin{subfigure}{0.32\textwidth}
    \centering
    \includegraphics[width=\linewidth]{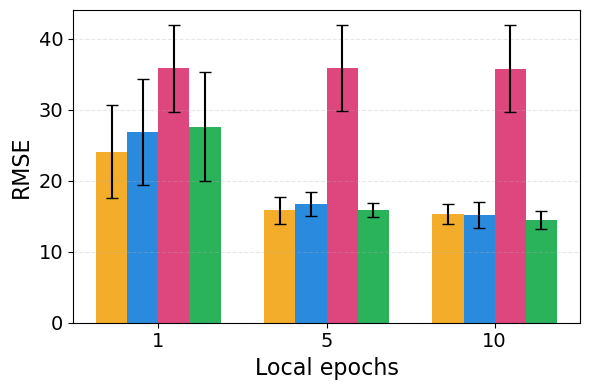}
    \caption{Task D}
  \end{subfigure}
  \caption{Average RMSE across all architectures for different numbers of local training epochs per round.}
  \label{fig:tasks_local_epochs}
\end{figure*}

\begin{figure*}[ht]
  \centering
  \begin{subfigure}{0.32\textwidth}
    \centering
    \includegraphics[width=\linewidth]{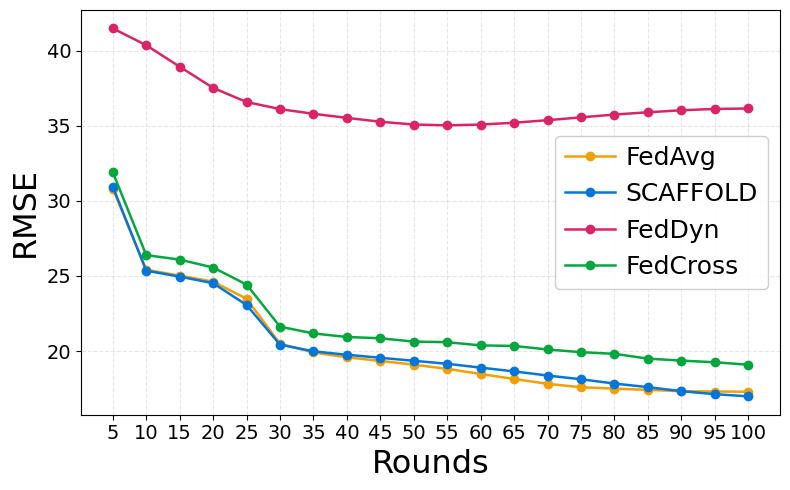}
    \caption{Task A}
  \end{subfigure}\hfill
  \begin{subfigure}{0.32\textwidth}
    \centering
    \includegraphics[width=\linewidth]{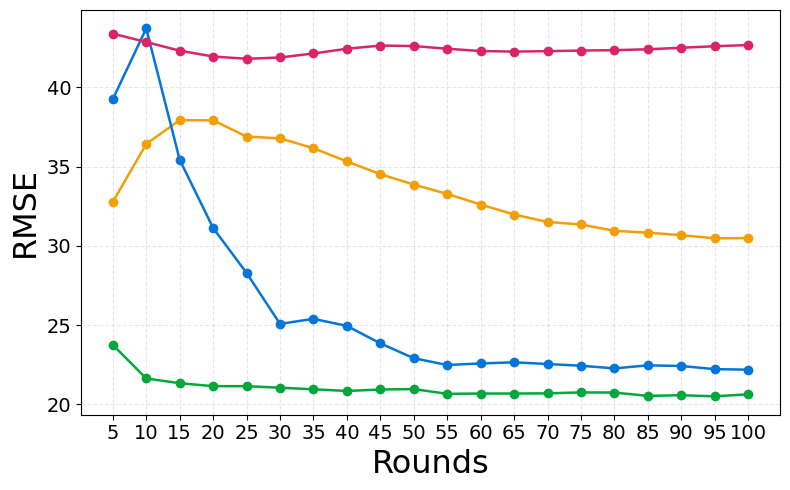}
    \caption{Task B}
  \end{subfigure}\hfill
  \begin{subfigure}{0.32\textwidth}
    \centering
    \includegraphics[width=\linewidth]{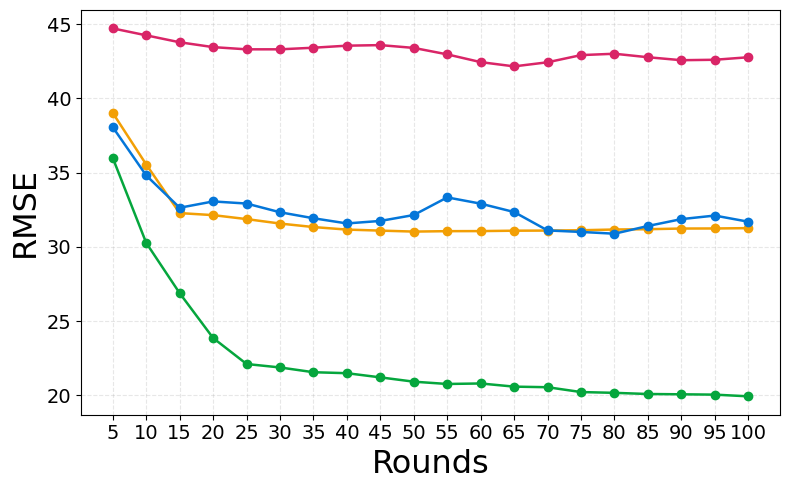}
    \caption{Task C}
  \end{subfigure}
  \caption{Average RMSE across all architectures over communication rounds, using 10 local epochs per round.}
  \label{fig:tasks_rounds}
\end{figure*}

\begin{figure*}[ht]
  \centering
  \begin{subfigure}{0.32\textwidth}
    \centering
    \includegraphics[width=\linewidth]{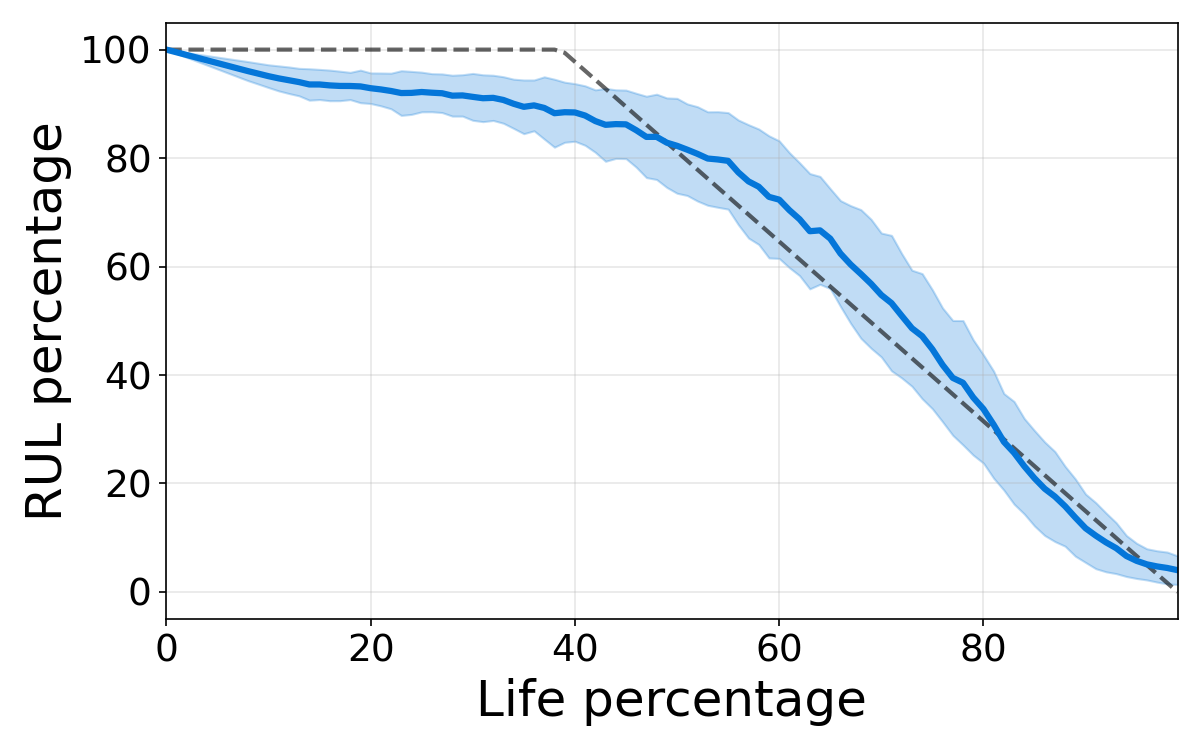}
    \caption{Task A: FedAvg using CNN}
  \end{subfigure}\hfill
  \begin{subfigure}{0.32\textwidth}
    \centering
    \includegraphics[width=\linewidth]{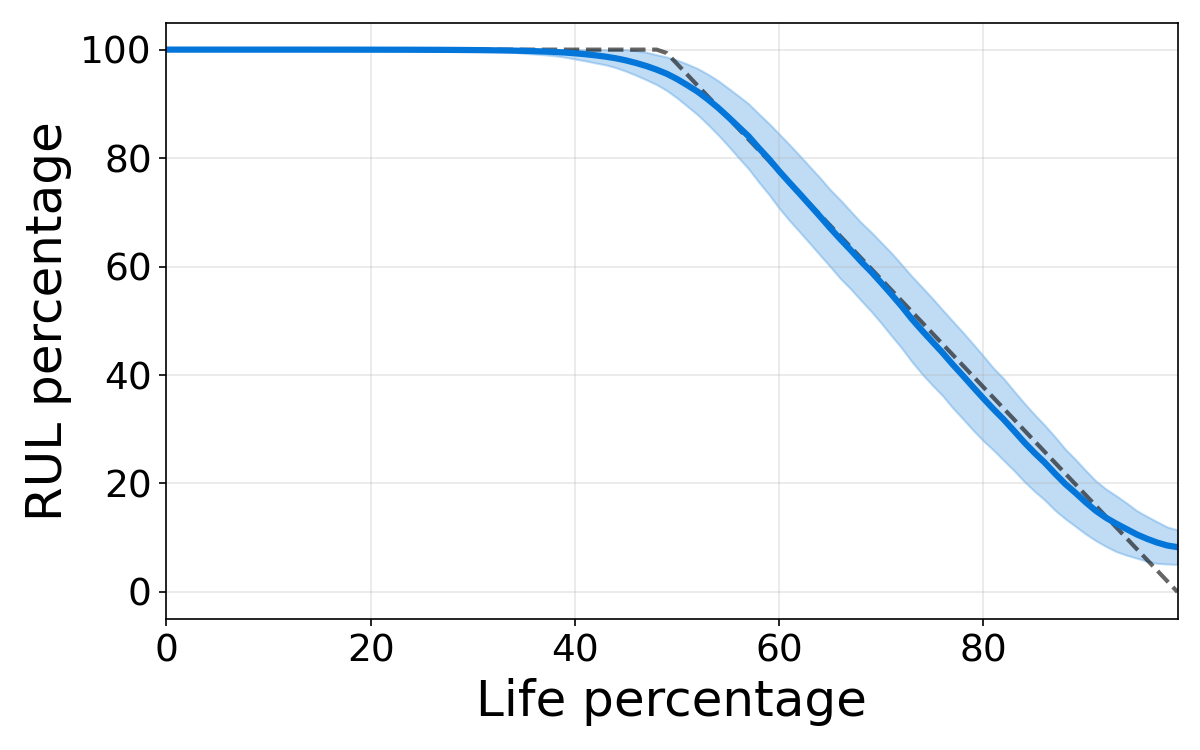}
    \caption{Task C: FedCross using RNN}
  \end{subfigure}
  \begin{subfigure}{0.32\textwidth}
    \centering
    \includegraphics[width=\linewidth]{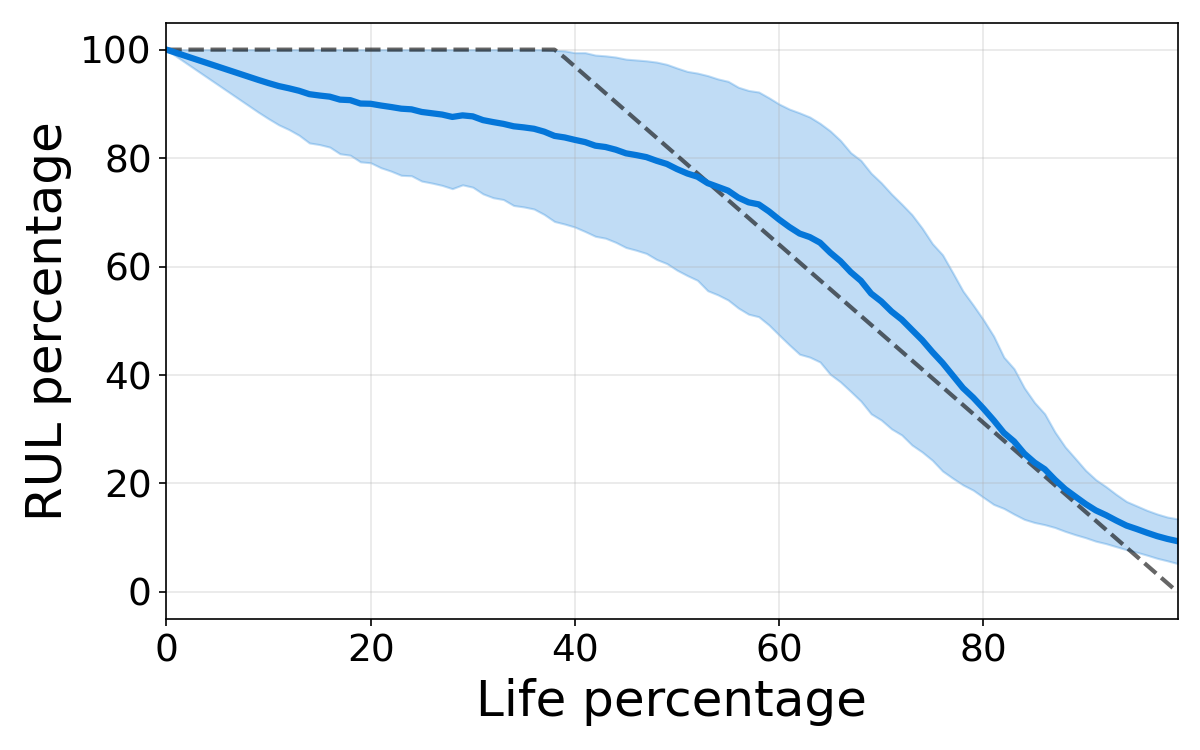}
    \caption{Task E: FedAvg using LSTM}
  \end{subfigure}
  \caption{Aggregated RUL trajectories over the life percentage.}
  \label{fig:tasks_ruls}
\end{figure*}

\subsection{Sensitivity to optimization hyperparameters}

A challenge in deploying federated solutions to edge environments lies in how to handle bandwidth limitations. Formally, the communication cost can be modeled as $C = 2RNP$, where $R$ denotes the total number of synchronization rounds, $N$ represents the number of clients participating in each round, and $P$ defines the total number of shared trainable parameters for a specific network architecture. Because the payload size $P$ and number of clients $N$ are fixed for a given experimental configuration, the communication overhead scales linearly with $R$. Therefore, deployment feasibility may be evaluated by analyzing the interaction between local training epochs and global synchronization frequency.

To this end, we conduct an analysis of the local optimization hyperparameters, reporting results averaged across model architectures for each task, shown in Fig.~\ref{fig:tasks_local_epochs} for tasks B, C and D. We first evaluate the effect of varying the number of local training epochs (1, 5, and 10) while keeping the total number of global communication rounds fixed at 100. In highly non-IID federated settings, performing multiple consecutive local gradient steps typically exacerbates client drift, pushing the local weights toward divergent, isolated minima. However, our findings indicate that the final generalization performance remains largely robust to the specific number of local epochs.

Building on this stability, we evaluate the impact of varying the total number of global rounds (from 10 to 100) while holding the local computation fixed to 10 epochs, to ensure that we capture long-training effects. 
As summarized in Fig.~\ref{fig:tasks_rounds}, 100 communication rounds are sufficient to reach stable performance; in some cases, notably for Task A, the trend seems to indicate further possible improvements;  a trade-off is however necessary when training in an edge-oriented regime.
On a negative note, FedDyn exhibits an almost-constant trend which questions its suitability to the tasks at hand. 

\subsection{Qualitative analysis of degradation trajectories}

Finally, we visualize common plots of actual RUL versus predicted RUL, to assess the models' capability to reliably estimate a component's lifespan during its operation. Fig.~\ref{fig:tasks_ruls} shows examples for a representative subset of the evaluated FL methods, models and tasks, visualizing the average predicted RUL, with confidence intervals computed across task splits. In the plots, it can be noticed that the length of the constant segment of the target RUL is not fixed across tasks. This is due to the target RUL clipping at 125, which has a different impact depending on the average length of samples in the task.

The degradation trajectories provide useful insight into the predictive uncertainty exhibited at different stages of a component lifecycle, which reflects the intrinsic difficulty of each task. As the degradation progresses toward the failure threshold, the fault signature becomes increasingly pronounced, leading to a narrowing of the error bars. Notably, there appears to be a tendency to slightly overestimate RUL as the device reaches the failure point. This should be further investigated and possibly addressed through ad-hoc loss terms that target end-of-life behavior.

\section{Conclusions}
\label{sec:conclusions}

Given the lack of consistent evaluation protocols and client partitioning schemes in current federated prognostics research, this paper introduces \textit{FedCMAPSS}, a standardized benchmark for RUL estimation in federated learning. Built on the widely used NASA C-MAPSS dataset, the benchmark defines five tasks spanning an IID baseline, several forms of statistical heterogeneity, and an extreme data-scarcity regime. Beyond task specification, FedCMAPSS provides a reproducible experimental protocol, including client-side preprocessing, windowing and target construction, a standardized training schedule, and multiple randomized splits. Together with publicly released code and data partitions, these elements enable fair and transparent comparisons across future methods.

Our experiments across architectures and FL methods yield practical insights. In the IID regime (Task~A), FedAvg performs strongly overall; under domain shift and label skew (Tasks~B-C), drift-mitigation strategies are generally more reliable. Feature skew (Task~D) is comparatively mild in our setup, whereas few-shot federation (Task~E) is the most challenging, and performance becomes more sensitive to the selected model architecture.

Further analysis indicates that final performance is relatively robust to the number of local epochs in the tested configurations. Moreover, convergence curves over communication rounds show that most methods reach a stable regime within 100 rounds, although in some cases additional rounds may be of benefit.

In future work, we plan to perform systematic hyperparameter optimization tailored to federated settings, extend the benchmark to additional FL paradigms, include outlier-robust evaluation criteria, and expand the suite with additional datasets.

\section*{Acknowledgment}
S. Palazzo and M. Pennisi acknowledge the project ECS4DRES, supported by the Chips Joint Undertaking under grant agreement number 101139790 and its members, including the top-up funding by Germany, Italy, Slovakia, Spain and The Netherlands.

\bibliographystyle{IEEEtran}
\bibliography{references}

@book{vachtsevanos2006intelligent,
  title={Intelligent fault diagnosis and prognosis for engineering systems},
  author={Vachtsevanos, George J and Lewis, Frank and Roemer, Michael and Hess, Andrew and Wu, Biqing and others},
  volume={456},
  year={2006},
  publisher={Wiley Online Library}
}

@article{zhang2019review,
  title={A review on deep learning applications in prognostics and health management},
  author={Zhang, Liangwei and Lin, Jing and Liu, Bin and Zhang, Zhicong and Yan, Xiaohui and Wei, Muheng},
  journal={IEEE Access},
  volume={7},
  pages={162415--162438},
  year={2019},
  publisher={IEEE}
}

@article{rezaeianjouybari2020deep,
  title={Deep learning for prognostics and health management: State of the art, challenges, and opportunities},
  author={Rezaeianjouybari, Behnoush and Shang, Yi},
  journal={Measurement},
  volume={163},
  pages={107929},
  year={2020},
  publisher={Elsevier}
}

@article{polverino2023machine,
  title={Machine learning for prognostics and health management of industrial mechanical systems and equipment: A systematic literature review},
  author={Polverino, Lorenzo and Abbate, Raffaele and Manco, Pasquale and Perfetto, Donato and Caputo, Francesco and Macchiaroli, Roberto and Caterino, Mario},
  journal={International Journal of Engineering Business Management},
  volume={15},
  pages={18479790231186848},
  year={2023},
  publisher={SAGE Publications Sage UK: London, England}
}

@article{qin2023dynamic,
  title={Dynamic weighted federated remaining useful life prediction approach for rotating machinery},
  author={Qin, Yi and Yang, Jiahong and Zhou, Jianghong and Pu, Huayan and Zhang, Xiangfeng and Mao, Yongfang},
  journal={Mechanical Systems and Signal Processing},
  volume={202},
  pages={110688},
  year={2023},
  publisher={Elsevier}
}

@article{arunan2023federated,
  title={A federated learning-based industrial health prognostics for heterogeneous edge devices using matched feature extraction},
  author={Arunan, Anushiya and Qin, Yan and Li, Xiaoli and Yuen, Chau},
  journal={IEEE Transactions on Automation Science and Engineering},
  volume={21},
  number={3},
  pages={3065--3079},
  year={2023},
  publisher={IEEE}
}

@inproceedings{mcmahan2017communication,
  title={Communication-efficient learning of deep networks from decentralized data},
  author={McMahan, Brendan and Moore, Eider and Ramage, Daniel and Hampson, Seth and y Arcas, Blaise Aguera},
  booktitle={Artificial intelligence and statistics},
  pages={1273--1282},
  year={2017},
  organization={PMLR}
}

@article{yang2019federated,
  title={Federated machine learning: Concept and applications},
  author={Yang, Qiang and Liu, Yang and Chen, Tianjian and Tong, Yongxin},
  journal={ACM Transactions on Intelligent Systems and Technology (TIST)},
  volume={10},
  number={2},
  pages={1--19},
  year={2019},
  publisher={ACM New York, NY, USA}
}

@article{nguyen2021federated,
  title={Federated learning for internet of things: A comprehensive survey},
  author={Nguyen, Dinh C and Ding, Ming and Pathirana, Pubudu N and Seneviratne, Aruna and Li, Jun and Poor, H Vincent},
  journal={IEEE communications surveys \& tutorials},
  volume={23},
  number={3},
  pages={1622--1658},
  year={2021},
  publisher={IEEE}
}

@article{chen2023bearing,
  title={Bearing remaining useful life prediction using federated learning with Taylor-expansion network pruning},
  author={Chen, Xi and Wang, Hui and Lu, Siliang and Yan, Ruqiang},
  journal={IEEE Transactions on Instrumentation and Measurement},
  volume={72},
  pages={1--10},
  year={2023},
  publisher={IEEE}
}

@inproceedings{bonawitz2017practical,
  title={Practical secure aggregation for privacy-preserving machine learning},
  author={Bonawitz, Keith and Ivanov, Vladimir and Kreuter, Ben and Marcedone, Antonio and McMahan, H Brendan and Patel, Sarvar and Ramage, Daniel and Segal, Aaron and Seth, Karn},
  booktitle={proceedings of the 2017 ACM SIGSAC Conference on Computer and Communications Security},
  pages={1175--1191},
  year={2017}
}

@article{zhu2024collaborative,
  title={Collaborative prognostics of lithium-ion batteries using federated learning with dynamic weighting and attention mechanism},
  author={Zhu, Rong and Peng, Weiwen and Ye, Zhi-Sheng and Xie, Min},
  journal={IEEE Transactions on Industrial Electronics},
  volume={72},
  number={1},
  pages={980--991},
  year={2024},
  publisher={IEEE}
}

@article{wang2020federated,
  title={Federated learning with matched averaging},
  author={Wang, Hongyi and Yurochkin, Mikhail and Sun, Yuekai and Papailiopoulos, Dimitris and Khazaeni, Yasaman},
  journal={arXiv preprint arXiv:2002.06440},
  year={2020}
}

@article{hanif2018comprehensive,
  title={A comprehensive review toward the state-of-the-art in failure and lifetime predictions of power electronic devices},
  author={Hanif, Abu and Yu, Yuechuan and DeVoto, Douglas and Khan, Faisal},
  journal={IEEE Transactions on Power Electronics},
  volume={34},
  number={5},
  pages={4729--4746},
  year={2018},
  publisher={IEEE}
}

@inproceedings{saxena2008damage,
  title={Damage propagation modeling for aircraft engine run-to-failure simulation},
  author={Saxena, Abhinav and Goebel, Kai and Simon, Don and Eklund, Neil},
  booktitle={2008 international conference on prognostics and health management},
  pages={1--9},
  year={2008},
  organization={IEEE}
}

@techreport{frederick2007user,
  title={User's guide for the commercial modular aero-propulsion system simulation (C-MAPSS)},
  author={Frederick, Dean K and DeCastro, Jonathan A and Litt, Jonathan S},
  year={2007}
}

@article{barbosa2025using,
  title={Using federated machine learning in predictive maintenance of jet engines},
  author={Barbosa, Asaph Matheus and Ngo, Thao Vy Nhat and Jafarigol, Elaheh and Trafalis, Theodore B and Ojoboh, Emuobosa P},
  journal={arXiv preprint arXiv:2502.05321},
  year={2025}
}

@inproceedings{rosero2020remaining,
  title={Remaining useful life estimation in aircraft components with federated learning},
  author={Rosero, Ra{\'u}l Homero Llasag and Silva, Catarina and Ribeiro, Bernardete},
  booktitle={PHM Society European Conference},
  volume={5},
  number={1},
  pages={9--9},
  year={2020}
}

@inproceedings{saxena2008metrics,
  title={Metrics for evaluating performance of prognostic techniques},
  author={Saxena, Abhinav and Celaya, Jose and Balaban, Edward and Goebel, Kai and Saha, Bhaskar and Saha, Sankalita and Schwabacher, Mark},
  booktitle={2008 international conference on prognostics and health management},
  pages={1--17},
  year={2008},
  organization={IEEE}
}

@inproceedings{byington2005verification,
  title={Verification and validation of diagnostic/prognostic algorithms},
  author={Byington, CS and Roemer, MJ and Kalgren, PW and Vachtsevanos, G},
  booktitle={Machinery Failure Prevention Technology Conference (MFPT 59)},
  year={2005}
}

@article{zhang2021federated,
  title={Federated learning for machinery fault diagnosis with dynamic validation and self-supervision},
  author={Zhang, Wei and Li, Xiang and Ma, Hui and Luo, Zhong and Li, Xu},
  journal={Knowledge-Based Systems},
  volume={213},
  pages={106679},
  year={2021},
  publisher={Elsevier}
}

@article{chen2023remaining,
  title={A remaining useful life estimation method based on long short-term memory and federated learning for electric vehicles in smart cities},
  author={Chen, Xuejiao and Chen, Zhaonan and Zhang, Mu and Wang, Zixuan and Liu, Minyao and Fu, Mengyi and Wang, Pan},
  journal={PeerJ Computer Science},
  volume={9},
  pages={e1652},
  year={2023},
  publisher={PeerJ Inc.}
}

@article{chen2023federated,
  title={Federated learning with network pruning and rebirth for remaining useful life prediction of engineering systems},
  author={Chen, Xi and Chen, Xinxian and Wang, Hui and Lu, Siliang and Yan, Ruqiang},
  journal={Manufacturing Letters},
  volume={35},
  pages={965--972},
  year={2023},
  publisher={Elsevier}
}

@inproceedings{hu2024fedcross,
  title={Fedcross: Towards accurate federated learning via multi-model cross-aggregation},
  author={Hu, Ming and Zhou, Peiheng and Yue, Zhihao and Ling, Zhiwei and Huang, Yihao and Li, Anran and Liu, Yang and Lian, Xiang and Chen, Mingsong},
  booktitle={2024 IEEE 40th International Conference on Data Engineering (ICDE)},
  pages={2137--2150},
  year={2024},
  organization={IEEE}
}

@inproceedings{acar2021federated,
  title={Federated Learning Based on Dynamic Regularization},
  author={Acar, Durmus Alp Emre and Zhao, Yue and Matas, Ramon and Mattina, Matthew and Whatmough, Paul and Saligrama, Venkatesh},
  booktitle={International Conference on Learning Representations},
  year={2021}
}

@article{li2020federated,
  title={Federated optimization in heterogeneous networks},
  author={Li, Tian and Sahu, Anit Kumar and Zaheer, Manzil and Sanjabi, Maziar and Talwalkar, Ameet and Smith, Virginia},
  journal={Proceedings of Machine learning and systems},
  volume={2},
  pages={429--450},
  year={2020}
}

@inproceedings{karimireddy2020scaffold,
  title={Scaffold: Stochastic controlled averaging for federated learning},
  author={Karimireddy, Sai Praneeth and Kale, Satyen and Mohri, Mehryar and Reddi, Sashank and Stich, Sebastian and Suresh, Ananda Theertha},
  booktitle={International conference on machine learning},
  pages={5132--5143},
  year={2020},
  organization={PMLR}
}

@article{liu2020deep,
  title={Deep anomaly detection for time-series data in industrial IoT: A communication-efficient on-device federated learning approach},
  author={Liu, Yi and Garg, Sahil and Nie, Jiangtian and Zhang, Yang and Xiong, Zehui and Kang, Jiawen and Hossain, M Shamim},
  journal={IEEE Internet of Things Journal},
  volume={8},
  number={8},
  pages={6348--6358},
  year={2020},
  publisher={IEEE}
}

@article{konevcny2016federated,
  title={Federated learning: Strategies for improving communication efficiency},
  author={Kone{\v{c}}n{\`y}, Jakub and McMahan, H Brendan and Yu, Felix X and Richt{\'a}rik, Peter and Suresh, Ananda Theertha and Bacon, Dave},
  journal={arXiv preprint arXiv:1610.05492},
  year={2016}
}

@inproceedings{sateesh2016deep,
  title={Deep convolutional neural network based regression approach for estimation of remaining useful life},
  author={Sateesh Babu, Giduthuri and Zhao, Peilin and Li, Xiao-Li},
  booktitle={International conference on database systems for advanced applications},
  pages={214--228},
  year={2016},
  organization={Springer}
}

@article{hochreiter1997long,
  title={Long short-term memory},
  author={Hochreiter, Sepp and Schmidhuber, J{\"u}rgen},
  journal={Neural computation},
  volume={9},
  number={8},
  pages={1735--1780},
  year={1997},
  publisher={MIT press}
}

@inproceedings{zheng2017long,
  title={Long short-term memory network for remaining useful life estimation},
  author={Zheng, Shuai and Ristovski, Kosta and Farahat, Ahmed and Gupta, Chetan},
  booktitle={2017 IEEE international conference on prognostics and health management (ICPHM)},
  pages={88--95},
  year={2017},
  organization={IEEE}
}

@article{vaccaro2023remaining,
  title={Remaining useful lifetime prediction of discrete power devices by means of artificial neural networks},
  author={Vaccaro, Alessandro and Biadene, Davide and Magnone, Paolo},
  journal={IEEE Open Journal of Power Electronics},
  volume={4},
  pages={978--986},
  year={2023},
  publisher={IEEE}
}

@article{chen2020machine,
  title={Machine remaining useful life prediction via an attention-based deep learning approach},
  author={Chen, Zhenghua and Wu, Min and Zhao, Rui and Guretno, Feri and Yan, Ruqiang and Li, Xiaoli},
  journal={IEEE Transactions on Industrial Electronics},
  volume={68},
  number={3},
  pages={2521--2531},
  year={2020},
  publisher={IEEE}
}

@article{lai2024fedcbe,
  title={FedCBE: A federated-learning-based collaborative battery estimation system with non-IID data},
  author={Lai, Rucong and Wang, Jie and Tian, Yong and Tian, Jindong},
  journal={Applied Energy},
  volume={368},
  pages={123534},
  year={2024},
  publisher={Elsevier}
}

@article{yilmaz2025federated,
  title={Federated learning-based state of charge estimation in electric vehicles using federated adaptive client momentum},
  author={Y{\i}lmaz, Metin and Yaz{\i}c{\i}, Ahmet and others},
  journal={IEEE Access},
  year={2025},
  publisher={IEEE}
}

@inproceedings{sun2024role,
  title={On the role of server momentum in federated learning},
  author={Sun, Jianhui and Wu, Xidong and Huang, Heng and Zhang, Aidong},
  booktitle={Proceedings of the AAAI Conference on Artificial Intelligence},
  volume={38},
  number={13},
  pages={15164--15172},
  year={2024}
}

@article{arivazhagan2019federated,
  title={Federated learning with personalization layers},
  author={Arivazhagan, Manoj Ghuhan and Aggarwal, Vinay and Singh, Aaditya Kumar and Choudhary, Sunav},
  journal={arXiv preprint arXiv:1912.00818},
  year={2019}
}

@article{soderkvist2024collaborative,
  title={Collaborative training of data-driven remaining useful life prediction models using federated learning},
  author={S{\"o}derkvist Vermelin, Wilhelm and Mishra, Madhav and Eng, Mattias P and Andersson, Dag and Kyprianidis, Konstantinos},
  journal={International Journal of Prognostics and Health Management},
  volume={15},
  number={2},
  year={2024}
}

@article{kamei2023comparison,
  title={A comparison study of centralized and decentralized federated learning approaches utilizing the transformer architecture for estimating remaining useful life},
  author={Kamei, Sayaka and Taghipour, Sharareh},
  journal={Reliability Engineering \& System Safety},
  volume={233},
  pages={109130},
  year={2023},
  publisher={Elsevier}
}

@inproceedings{peel2008data,
  title={Data driven prognostics using a Kalman filter ensemble of neural network models},
  author={Peel, Leto},
  booktitle={2008 international conference on prognostics and health management},
  pages={1--6},
  year={2008},
  organization={IEEE}
}

@inproceedings{pennisi2022gan,
  title={Gan latent space manipulation and aggregation for federated learning in medical imaging},
  author={Pennisi, Matteo and Proietto Salanitri, Federica and Palazzo, Simone and Pino, Carmelo and Rundo, Francesco and Giordano, Daniela and Spampinato, Concetto},
  booktitle={International Workshop on Distributed, Collaborative, and Federated Learning},
  pages={68--78},
  year={2022},
  organization={Springer}
}

@inproceedings{pennisi2023experience,
  title={Experience replay as an effective strategy for optimizing decentralized federated learning},
  author={Pennisi, Matteo and Salanitri, Federica Proietto and Bellitto, Giovanni and Spampinato, Concetto and Palazzo, Simone and Casella, Bruno and Aldinucci, Marco},
  booktitle={Proceedings of the IEEE/CVF International Conference on Computer Vision},
  pages={3376--3383},
  year={2023}
}

@inproceedings{mineo2023fedetr,
  title={FeDETR: A Federated Approach for Stenosis Detection in Coronary Angiography},
  author={Mineo, Raffaele and Sorrenti, Amelia and Proietto Salanitri, Federica},
  booktitle={International Conference on Image Analysis and Processing},
  pages={189--200},
  year={2023},
  organization={Springer}
}


\end{document}